\documentclass{article}

\usepackage[nonatbib, preprint]{neurips_2019}

\usepackage[T1]{fontenc}    
\usepackage{hyperref}       
\usepackage{url}            
\usepackage{booktabs}       
\usepackage{amsfonts}       
\usepackage{nicefrac}       
\usepackage{microtype}      

\title{Distance Assessment and Hypothesis Testing \\of High-Dimensional Samples using \\Variational Autoencoders} 

\author{%
  Marco Henrique de Almeida In\'acio \\
  University of S\~{a}o Paulo\\
  University of S\~{a}o Carlos \\	
Budapest University of Technology and Economics \\
  \texttt{m@marcoinacio.com} \\
  \And
  Rafael Izbicki \\
  University of S\~{a}o Carlos \\
  \texttt{rafaelizbicki@gmail.com} \\
  \AND
  B\'alint Gyires-T\'oth \\
  Budapest University of Technology and Economics \\
  \texttt{toth.b@tmit.bme.hu} \\
}

\usepackage{amsfonts}
\usepackage{lmodern}
\usepackage{longtable}
\usepackage{float} 
\usepackage{setspace}                   
\usepackage{indentfirst}                
\usepackage{makeidx}                    
\usepackage[nottoc]{tocbibind}          
\usepackage{courier}                    
\usepackage{type1cm}                    
\usepackage{listings}                   
\usepackage{titletoc}
\usepackage{multirow}
\usepackage{listings}
\usepackage{hyperref}
\hypersetup{
    colorlinks,
    citecolor=black,
    filecolor=black,
    linkcolor=black,
    urlcolor=black
}

\usepackage{algorithm}
\usepackage{algpseudocode}

\usepackage{xpatch}
\xpatchcmd{\algorithmic}{\setcounter}{\algorithmicfont\setcounter}{}{}
\providecommand{\algorithmicfont}{}

\usepackage[font=small,format=plain,labelfont=bf,up,textfont=it,justification=justified,singlelinecheck=false]{caption}
\usepackage[usenames,svgnames,dvipsnames]{xcolor}

\usepackage{enumitem} 
\usepackage{graphicx} 
\usepackage{amsmath} 
\usepackage{multicol}

\usepackage{placeins} 
\usepackage{savefnmark}
\usepackage{caption}
\usepackage{multicol}
\usepackage{subcaption}
\usepackage{booktabs}
\usepackage{wrapfig}

\usepackage{array}
\newcolumntype{H}{>{\setbox0=\hbox\bgroup}c<{\egroup}@{}}

\usepackage{listings}
\usepackage[backend=biber,style=ieee,natbib=true,doi=true,url=false]{biblatex}
\usepackage{enumitem}
\setlist{nosep} 
\setlist{itemsep=1pt, topsep=3pt}
\usepackage{algorithm}
\usepackage{algpseudocode}

\begin{document}

\maketitle

\begin{abstract}
Given two distinct datasets, an important question 
is if they have arisen from the the same data generating function or alternatively how their data generating functions diverge from one another. 
%
In this paper, we introduce an approach for measuring the distance between two datasets with high dimensionality using variational autoencoders. This approach is augmented by a permutation hypothesis test in order to check the hypothesis that the data generating distributions are the same within a significance level. 
We evaluate both the distance measurement and hypothesis testing approaches on generated and on public datasets.
%
%
According to the results the proposed approach can be used for data exploration (e.g. by quantifying the discrepancy/separability between categories of images), which can be particularly useful in the early phases of the pipeline of most machine learning projects.
\end{abstract}

\section{Introduction}
An important problem in the areas of machine learning and Statistics, is question of whether distinct samples (or datasets) arise a single data generating function (see \citep{gretton2012kernel,holmes2015two, soriano2015bayesian}, for instance) and if not, by how much suuch data generating functions diverge from one another. Such information could be crucial in most machine learning projects such as comparisons of datasets of images from multiple categories (e.g.: aiming to distinguish photos of dogs from the photos of cats).
Given that,
we present a novel approach to measure the distance between two high-dimensional datasets using variational autoencoders \citep{vae}. We additionally propose a hypothesis test in order to check the hypothesis the data generating distributions are the same.

The remaining of this paper is organized as follows:
in Section \ref{sec:vae_related_work}, we present a brief description of work related to our proposed method.
In Section \ref{sec:vae}, we present an introduction to variational inference and variational autoencoders, and show that the latter can be interpreted as density estimation procedure.
In Section \ref{sec:two_sample_comparison}, we explain how variational autoencoders can be used as a method of two sample comparison.
In Section \ref{sec:compare}, our method of populations distance measurement is presented while in
\ref{sec:htest}, we show how this method can be adapted to hypothesis testing procedure. Both sections also show applications of the methods to simulated and real-world datasets.
Finally, section \ref{sec:conclusion} concludes article with final remarks.
Additionally, in Appendix \ref{sec:configs_and_software}, we present the configurations of the software and neural networks used and a link to our implementation as well.

\section{Related work}
\label{sec:vae_related_work}
Ours approaches build upon the work of \citep{deAlmeidaIncio2018}, which gives a framework for obtaining an uncertainty measure of the distance between populations for a given  density estimation method and a metric in the space of density functions. Nonetheless, the default method proposed in that work lacks scalability to large datasets (given the impossibility of data subsampling of MCMC methods \citep{betancourt_mcmc_subsampling}) and high-dimensional data. In this work, we overcome this issues by using variational autoencoders as the density estimation method and of a distinct metric which has an analytic solution for high-dimensional datasets.

As the additional most closely related works, we can highligh the following ones which deal with the question of two-sample testing:
\citep{gretton2012kernel}, which proposes a two-sample test comparison using reproducing kernel Hilbert space theory;
\citep{holmes2015two} and \citep{soriano2015bayesian}, which propose and analyse novel Bayesian methods for two-sample comparison.

On the side of the closely related problem of independence testing, we could emphasize: \citep{pfister2016kernel}, which proposes a method for two-sample independence testing based on the two variable Hilbert-Schmidt independence criterion; \citep{ramdas2017wasserstein}, which presents a short survey on Wasserstein distance based test statistics for two sample independence testing; and \citep{1908.00105} which proposes a novel conditional independence testing method which is then compared to other established methods.

Also, as other works in areas relevant to the subject, we highlight 
\citet{1905.00414}, which proposes a new method of neural networks representation comparison (as well as review previous ones): this is related to the question of datasets comparison since such a neural networks comparison method can be applied to neural networks trained on distinct datasets;  \citet{1512.09300}, which proposes a variant of the VAE that better measure similarities in
data space (when compared to the vanilla VAE), with the caveat that it is not used for databases comparison; \citet{an2015variational}, which uses VAEs for anomaly detection: in this case however, it diverges from our objective in the sense that it is done at a single instance level, instead of datasets level; and finally, \citet{1280752}, which evaluate existing similarity measurement methods in the context of image retrieval.

\section{Variational Autoencoders}
\label{sec:vae}
In this section, we give a brief of review the most essential parts of the variational autoencoders framework\citet{vaeoriginal} relative to its usage in our proposed methods (which is discussed in the subsequent sections of the article).

In this sense, given an i.i.d. random sample $D = (X_1, X_2, ..., X_n)$,  consider the problem of estimating the probability density of $X_i$. The solution given by variational autoencoders is to encode the information of each $X_i$ in latent random variables $Z = (Z_1, Z_2, ..., Z_n)$ which are linked by a parameter $\theta$.

In principle, inference on such model could be carried with the standard Bayesian framework:
\begin{align*}
P(\theta,Z|D=d) = \frac{P(D=d|Z, \theta) P(Z| \theta) P(\theta)}{P(D=d)}
= \frac{P(\theta) \prod_{i=1}^n P(X_i=x_i|Z_i, \theta) P(Z_i)}{P(D=d)}
\end{align*}
However, instead it is carried out\footnote{A full Bayesian framework would be prohibitively slow in this case, except for small artificial neural networks (or using simpler functions instead). On the other hand ordinary (non-Bayesian) neural works (i.e.: maximization of evidence) should be able to produce a model rich enough to capture a large class of density functions.} by maximizing the evidence $P(D=d; \theta) = P_\theta(D=d)$ (i.e.: maximum likelihood estimation or maximum a posteriori assuming we had uniform prior on $\theta$ for this parameterization). Additionally consider the following structure:
$$P_\theta(D=d|Z) = \prod_{i=1}^n \mathcal{N}(X_i=x_i; (\mu_i, \sigma_i)=g_\theta(Z_i)) \mbox{\ \ \ and \ \ } Z \sim \mathcal{N}(0,1),$$
where $g_\theta$ is a complex function with parameter $\theta$. Note that, if $g_\theta$ is complex enough, we can actually model any distribution of $X_i=x_i|Z_i$ \citep{devroye1986sample}. So, we use artificial neural networks due to its flexibility, scalability and the universal approximation theorem \citep{Hornik1989}. As explained in \citep{tutorial_vae}, this problem cannot be directly solved due to the curse of dimensionality.

\subsection{Variational inference}
The curse of dimensionality can be solved using variational inference which consists of optimizing
\begin{align*}
~
&
~ \log P_\theta(D=d) - \mathbf{D}_{KL}(Q_\phi^{(Z|D=d)} | P_\theta^{(Z|D=d)})
\\
~
&
= E_{Q_\phi} \left [\log P_\theta(D|Z) \right | D=d] - \mathbf{D}_{KL}(Q_\phi^{(Z|D=d)} | P^{(Z)})
\end{align*}
\begin{figure}[!htb]
  \centering
  \includegraphics[width=0.66\textwidth]{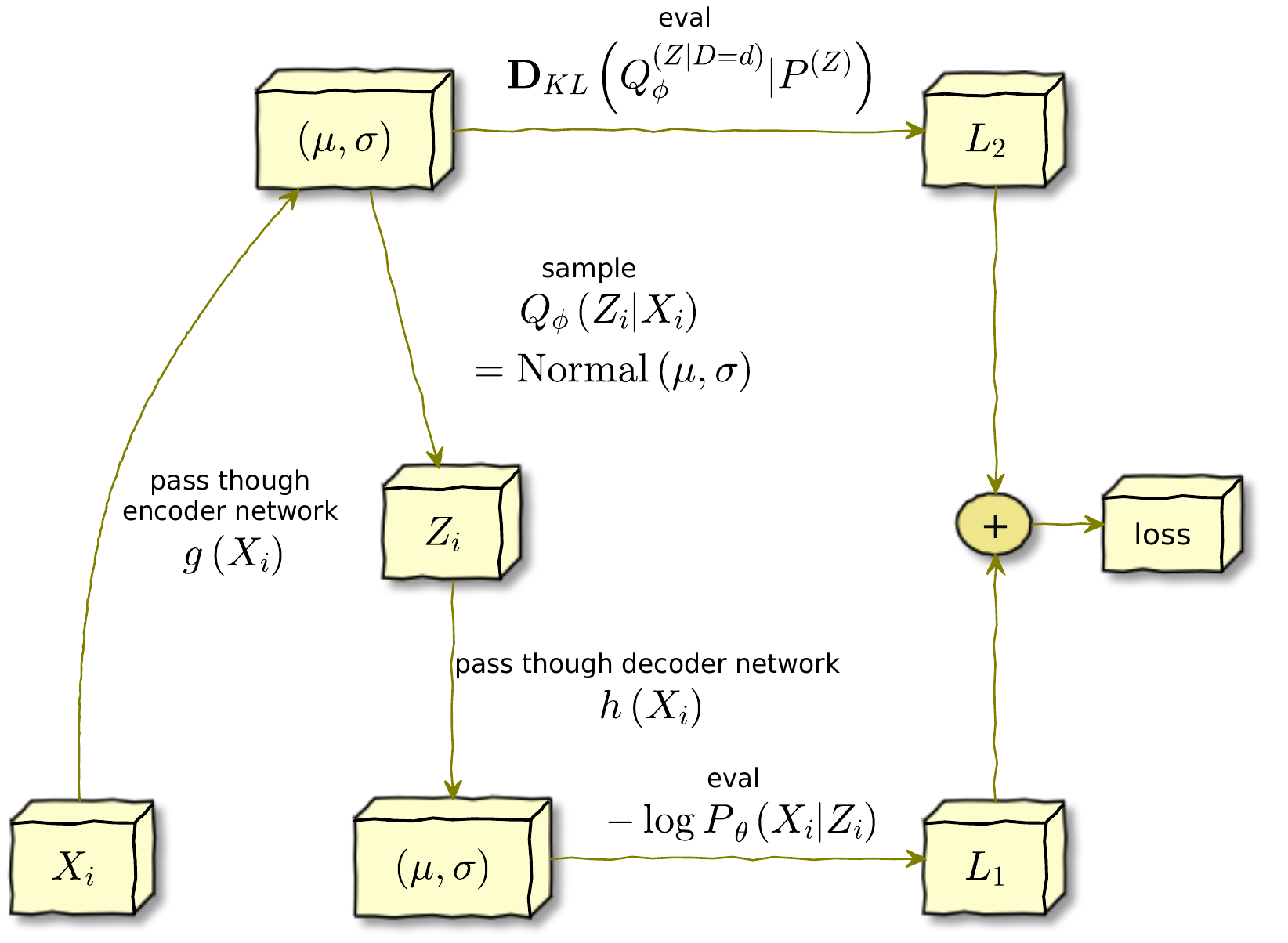}
  \caption{VAE training procedure.}
  \label{fig:diagram_vae}
\end{figure}
given the right choice of $Q$:
$$Q_\phi(Z_i|X_i=x_i) = \mathcal{N}(Z_i; (\mu_i, \sigma_i)=h_\phi(x_i))$$
This framework is called variational autoenconder \citep{vae} and it solves the curse of dimensionality.
The training procedure for variational autoencoders is presented in Figure \ref{fig:diagram_vae}.

\subsection{Generative models}
The question that remains is how to generate new instances $\widetilde{X}_j$ from the trained model. This can be done by applying
$
P_{\hat\theta}(\widetilde{X}_j) = \int P_{\hat\theta}(\widetilde{X}_j|\widetilde{Z_j}=z) P^{(\widetilde{Z_j})}(\mathrm{d}z)
$
, i.e.: sample from the prior of $Z$, apply it on the neural network $g_\theta$ and then sample from a $\mathcal{N}((\mu,\sigma)=g_\theta)$. Therefore, variational autoencoders can be seen in this sense as a Gaussian\footnote{Note that variational autoencoders setup can be used with distributions other then Gaussians; e.g.: discrete data with Bernoulli distribution.} mixture model (of an infinite number of Gaussians), and therefore, a density estimator.

\subsection{Identifiability of the mixture of Gaussians}
\label{sec:identifiability}
As per the structure of variational autoencoders, the distribution of such mixture of Gausssians is not per se identifiable: if we train a variational autoencoders framework on a dataset, we will get a ``generator'' of pairs $(\mu, \sigma)$ which we can call $m_1$, and if we train a variational autoencoders framework with identical structure on the same dataset, we will get another ``generator'' of pairs $(\mu, \sigma)$ which we can call $m_2$.

Generators $m_1$ and $m_2$ are not ensured to gives us the same distribution over samples of pairs $(\mu,\sigma)$. Nonetheless, the induced final density (i.e., the Gaussian mixture) should be same analytically (i.e.: ignoring the stochastic variation that estimation methods induce, specially SGDs).

\section{Two sample comparison: definition of the criterion}
\label{sec:two_sample_comparison}

Given two datasets $D_1$ and $D_2$, we can obtain via Bayesian inference the posterior PDF estimation for each of them, $P(f_1 | D_1)$ and $P(f_2 | D_2)$, generate samples $f_1$ and $f_2$ from each posterior distribution, and then calculate of the distance between these generated pairs of $f_1$ and $f_2$ using the integrated squared distance:
$$
\int (f_1(x) - f_2(x))^2 \mathrm{d}x,
$$
therefore obtaining a posterior sample of the distance between the PDFs\footnote{See \citet{deAlmeidaIncio2018} for details on this procedure.}.

One of the drawbacks of the proposed method, however, is the computational time that would be required to obtain an integrated distance metric for high-dimensional datasets. To tackle this issue, we propose here the usage new instances of a variational autoencoders and the use of Kullback-Leibler divergence as the distance measurement of choice due to the fact that it has a simple analytic solution for the case two $d$-dimensional Gaussian multivariate distributions \citep{divergence-two-gaussians}:
\begin{align*}
& \mathbf{D}_{KL}(\mathcal{N}(\mu_1, \Sigma_1), \mathcal{N}(\mu_2, \Sigma_2)) =
\\
& \frac{1}{2}
\left[\log|\Sigma_2| - \log |\Sigma_1|
- d
+ \text{tr} \{ \Sigma_2^{-1}\Sigma_1 \}
+ (\mu_2 - \mu_1)^T \Sigma_2^{-1}(\mu_2 - \mu_1)
\right]
\end{align*}
Notice that an additional computational simplification is possible given that in the proposed formulation of variational autoencoders the covariance matrix of each Gaussian is diagonal:
\begin{align*}
& \mathbf{D}_{KL}(\mathcal{N}(\mu_1, \sigma_1^T I), \mathcal{N}(\mu_2, \sigma_2^T I)) =
\\
& \frac{1}{2} \left[
2 \left(\sum_{i=1}^d \log \sigma_{2,i}  - \log \sigma_{1,i}\right)
- d
+ \left(\sum_{i=1}^d \sigma_{1,i}^2 / \sigma_{2,i}^2\right)
+ \left(\sum_{i=1}^d \sigma_{2,i}^2 (\mu_{2,i} - \mu_{1,i})^2\right)
\right]
\end{align*}
Moreover, in case of images, as it is customary on the VAE literature (see \citep{vae, tutorial_vae}, for instance), we work with multi-dimensional Bernoulli distributions (with dimensions independent from each other). In this case, the Kullback-Leibler can also be obtained easily and we have:
\begin{align*}
& \mathbf{D}_{KL}(\text{Bernoulli}(p), \text{Bernoulli}(q)) + \mathbf{D}_{KL}(\text{Bernoulli}(q), \text{Bernoulli}(p))
\\ & =
\sum_{i=1}^d
(q_i - p_i) (\log(q_i) - \log(p_i) + \log(1 - p_i) - \log(1 - q_i))
\end{align*}

However, there are two caveats in using variational autoencoders instead of Bayesian posterior samples. First, that variational autoencoders new instance samples cannot be interpreted as Bayesian samples. Still, it is possible to interpret the mean of the distance's samples as a measure of distance. Another problem is the one mentioned in Section \ref{sec:identifiability} regarding the identifiability of the samples of $(\mu, \sigma)$. To tackle this issue, we trained the variational autoencoders multiple times (we call these ``refits'') for each dataset (i.e. using distinct initialization seeds for the network parameters) and used the new instances pairs $(\mu, \sigma)$ from each of them in equal proportion. Yet another, is the fact that the Kullback-Leibler divergence is not symmetric and the dimensionallity of the dataset (i.e.: number of features) affects its magnitude, we solve this by working with the average symmetric KL divergence: 
$$
\mathbb{D}(x, y) =
\frac{
\mathbf{D}_{KL}(P_x, P_y) + \mathbf{D}_{KL}(P_y, P_x)
}
{2 d}
$$
We present the procedure in Algorithm \ref{alg:comparison}.

\begin{algorithm}
 \caption{ \small Generating divergence samples}\label{alg:comparison}
 \textbf{Input:} {\small  
 dataset $D_1$,
 dataset $D_2$,
 number of desired samples per refit $n$,
 number of desired refits $R$
 } \\
 \textbf{Output:} {\small  
 divergence samples $S$.
 } 
 \begin{algorithmic}[1]
  \For{$i \in \{1,\ldots,k\}$}
      \State Train VAE $V_1$ from $D_1$
      \State Train VAE $V_2$ from $D_2$
      \For{$j \in \{1,\ldots,n\}$}
          \State Generate a sample $s_1$ from $V_1$ (e.g.: a pair $(\mu, \sigma)$ for Gaussian VAE).
          \State Generate a sample $s_2$ from $V_2$.
          \State Calculate $\mathbb{D}(s_1, s_2)$ and store it on $S$. 
      \EndFor
  \EndFor
 \end{algorithmic}
\end{algorithm}

\section{Measuring the comparison criterion}
\label{sec:compare}
In Section \ref{sec:two_sample_comparison}, we defined a method to measure the distance between dataset distributions. A yardstick (measurement instrument) is still required in order to say what is a ``low''  and ``high'' distances.
In this sense, in similar fashion to \citet{deAlmeidaIncio2018}, we work with the divergence between two known distribution as the baseline to interpret such measure.

Therefore, in this respect, for Gaussian VAEs we can work with
$
\mathbb{D}(\mathcal{N}_0, \mathcal{N}_1)
$,
where
$\mathcal{N}_0$ is a multivariate Gaussian with covariance given by an identity matrix and mean given by a vector of zeros
and 
$\mathcal{N}_1$ is a multivariate Gaussian with covariance given by an identity matrix and mean given by a vector of ones. Note that $\mathbb{D}(\mathcal{N}_0, \mathcal{N}_1) = 1/2$.

For binomial VAEs, a possibility is to use known binomial distributions as the baseline as done in the Section \ref{subsec:eval_compare_cifar10}.

Note that from the perspective of applying this method to images, it can also be interpreted as a data exploration tool, as it helps exploring the separability and uncertainty of classes of images and the relation to their data generating processing, as we shall see in the next subsection.

\subsection{Evaluation (images)}
\label{subsec:eval_compare_cifar10}
The proposed method of distance measurement was applied to CIFAR10 data\citep{cifar10} using the VAE as generator of binomial distributions. The dataset consists of images from 10 distinct categories, with each category containing 5000 images. To make the comparison fair when comparing a category to itself and when comparing a category to another, we chose to work with half of each category dataset (2500 images) to train each VAE; i.e.: when comparing category 1 to category 2, one VAE is trained with 2500 images from category 1 and the other is trained with 2500 images from category 2; on the other hand, when comparing category 1 to itself, each VAE is trained with half (2500 images) of the category 1 dataset. We worked with 90 VAE refits for each dataset.

\begin{figure}[!htb]
\begin{subfigure}{.5\textwidth}
  \centering
  \includegraphics[width=\linewidth]{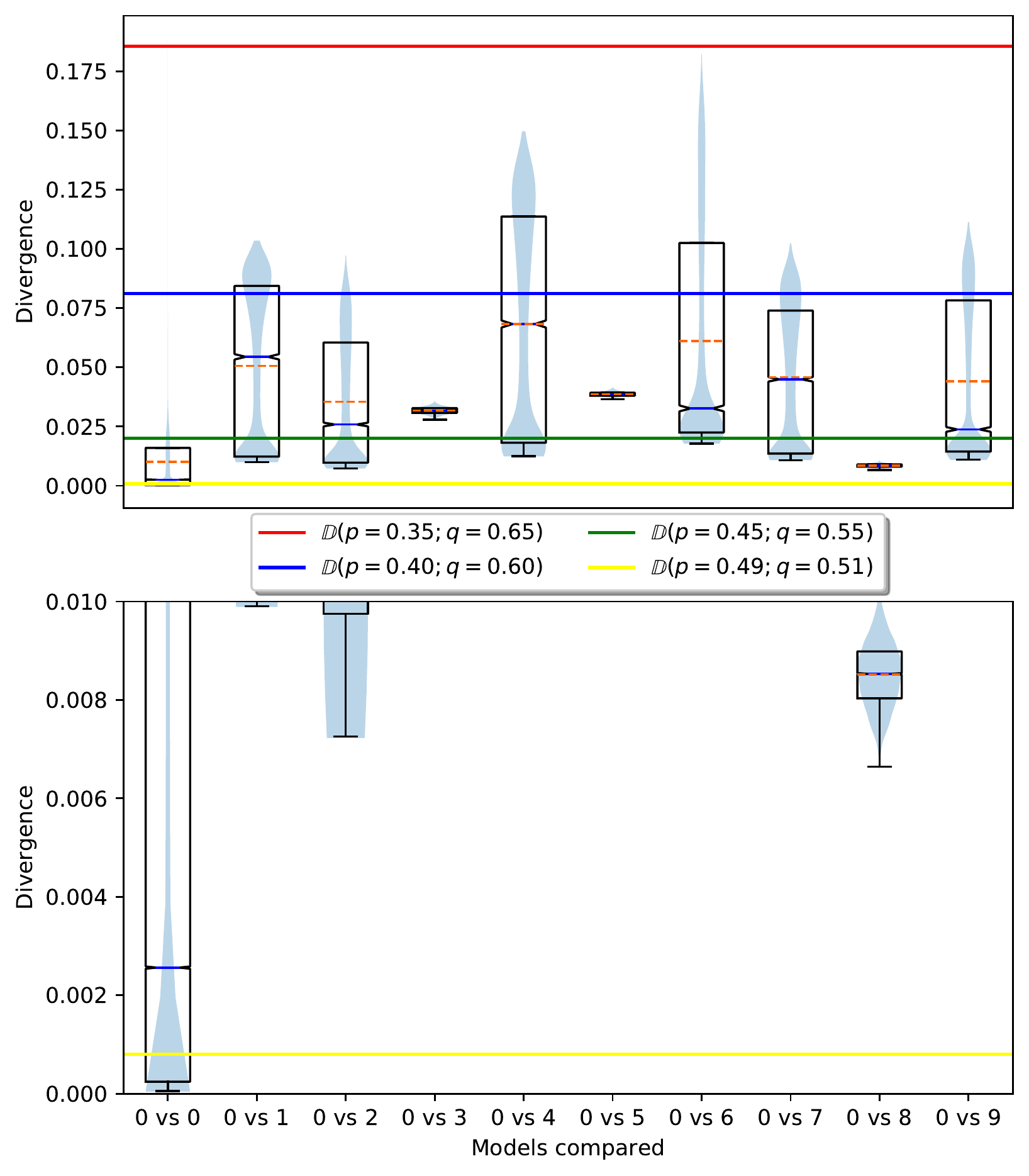}
\end{subfigure}%
\begin{subfigure}{.5\textwidth}
  \centering
  \includegraphics[width=\linewidth]{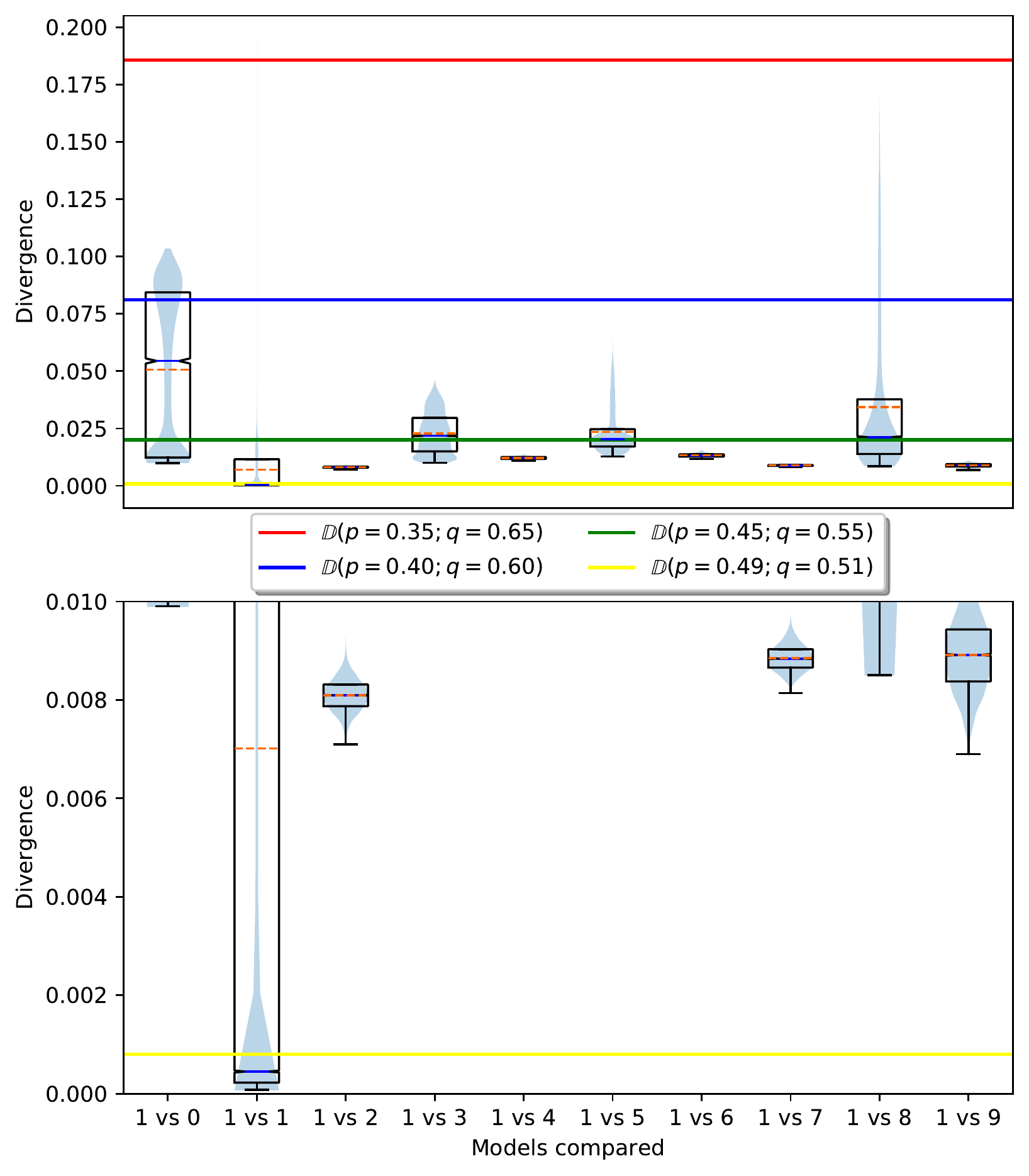}
\end{subfigure}
\begin{subfigure}{.5\textwidth}
  \centering
  \includegraphics[width=\linewidth]{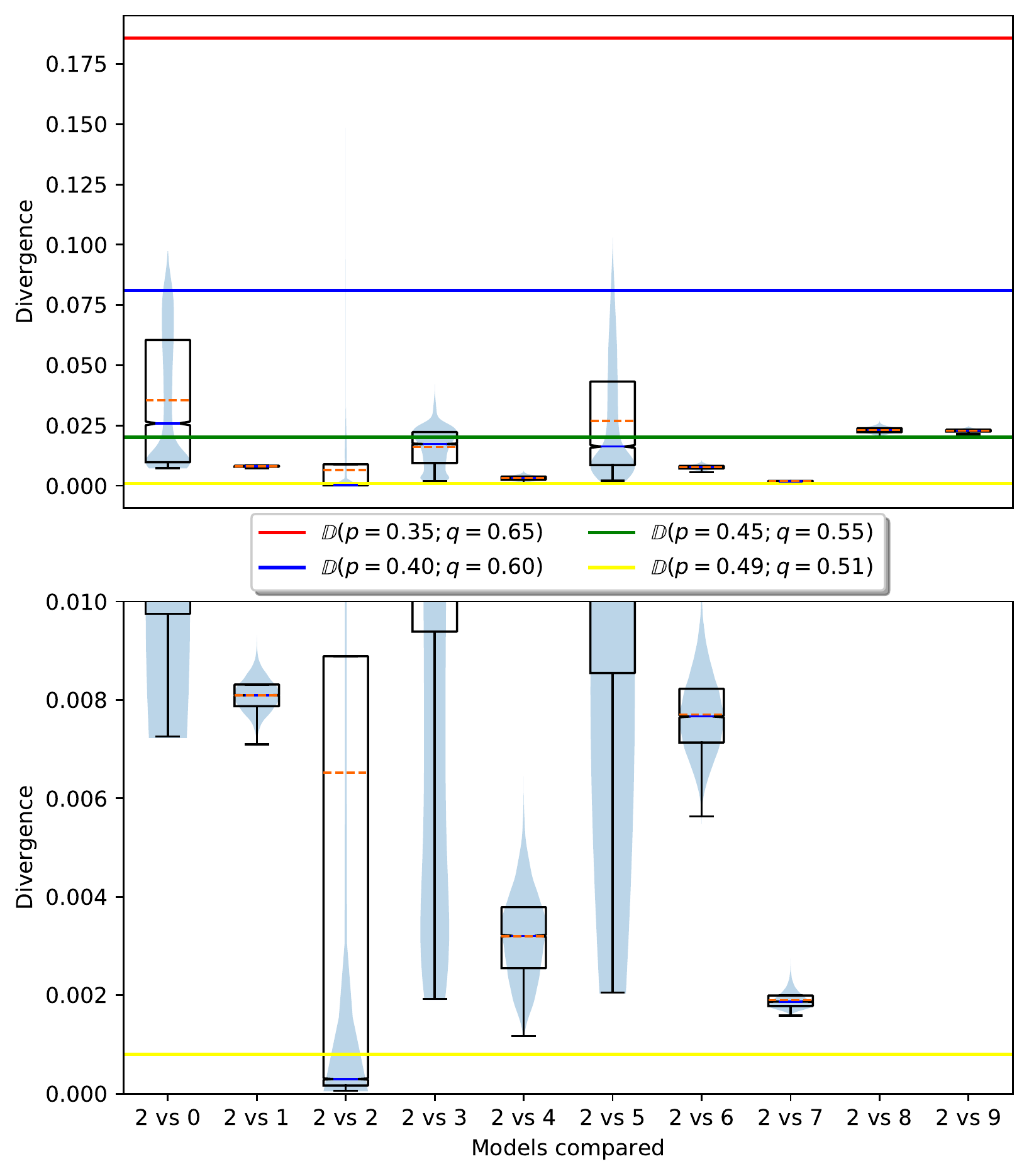}
\end{subfigure}%
\begin{subfigure}{.5\textwidth}
  \centering
  \includegraphics[width=\linewidth]{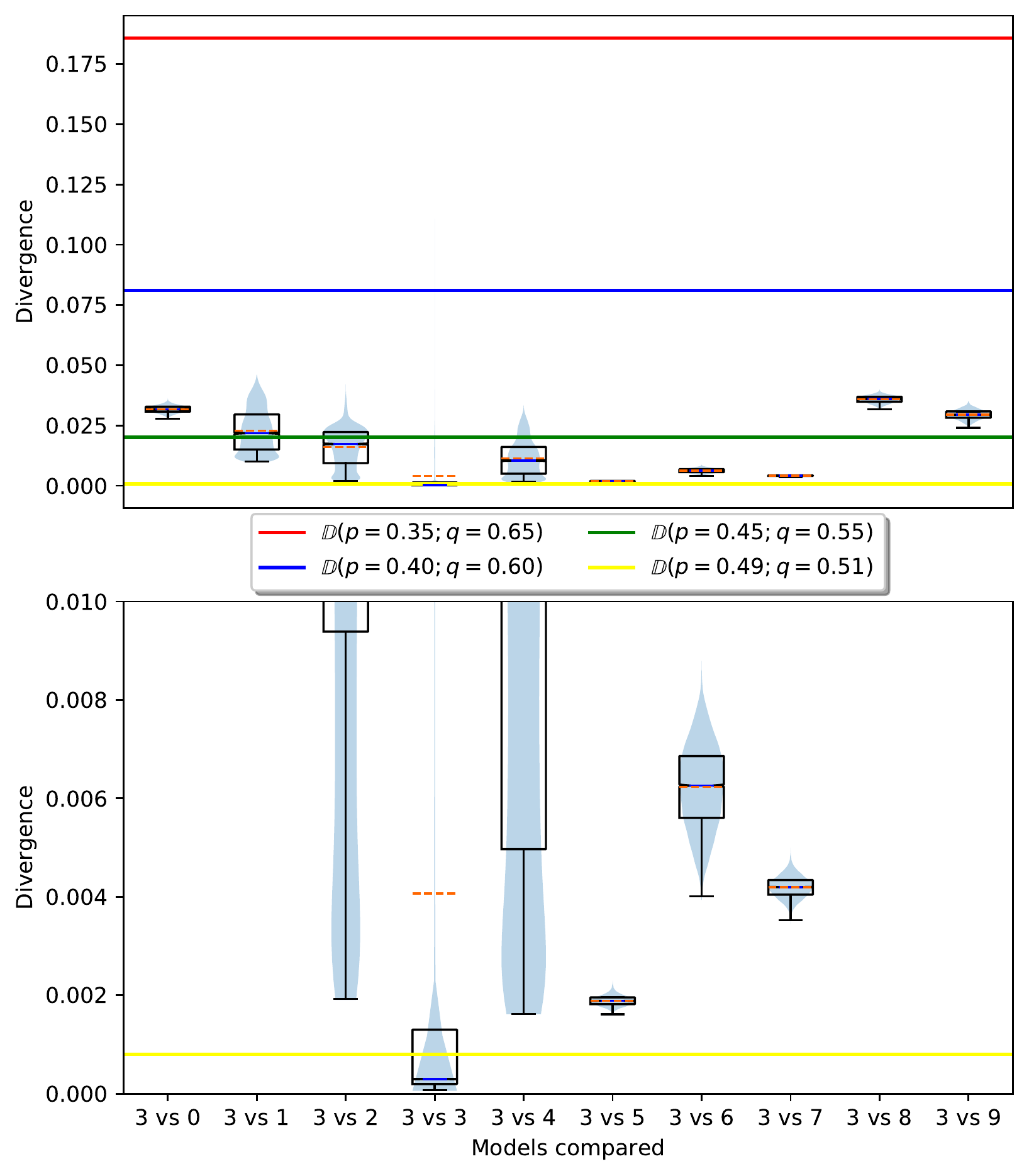}
\end{subfigure}
\caption{Box plots of samples from our divergences comparing categories 0 to 3 to all categories.}
\label{fig:cifar_compare_1}
\end{figure}

\begin{figure}[!htb]
\begin{subfigure}{.5\textwidth}
  \centering
  \includegraphics[width=\linewidth]{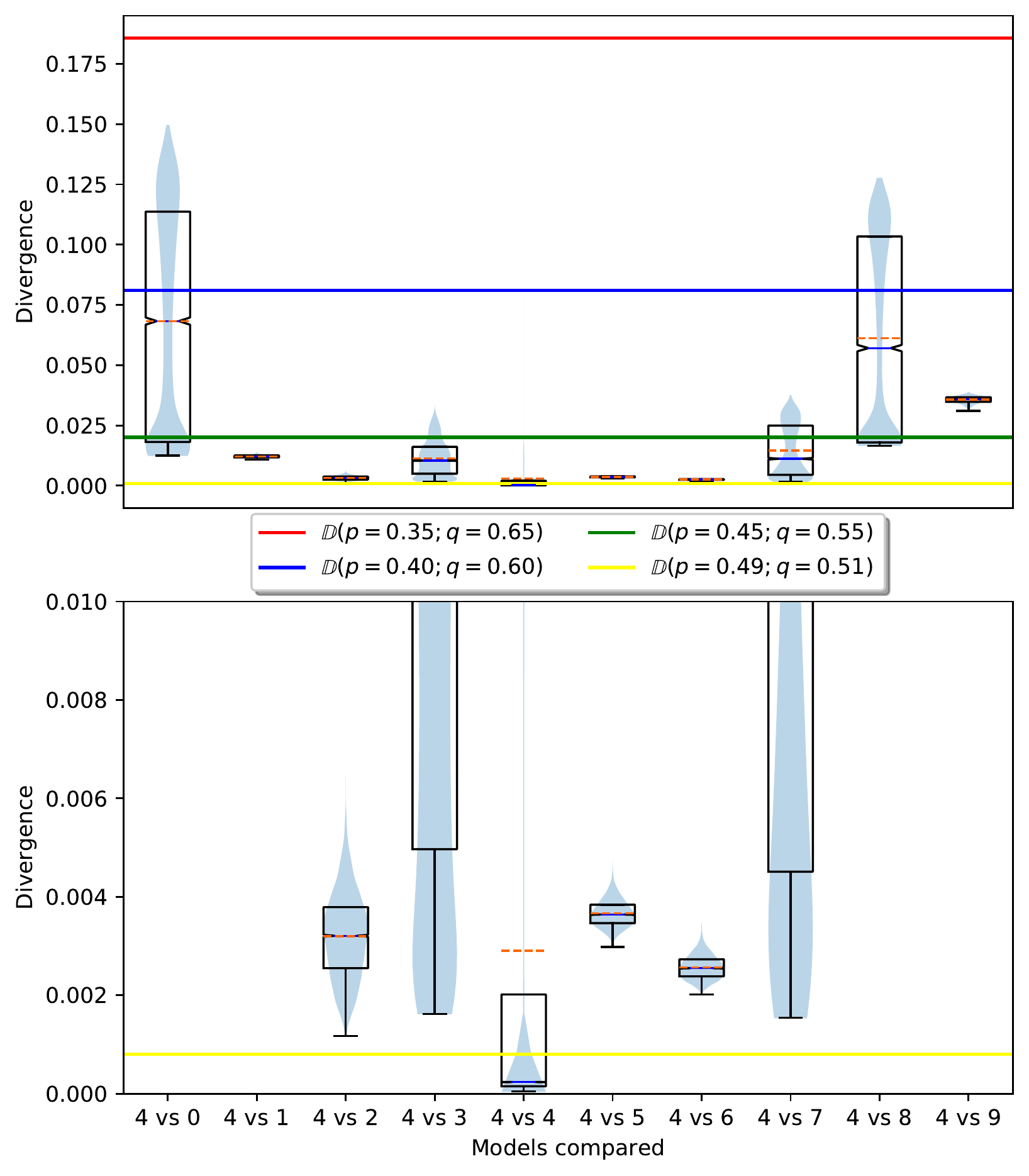}
\end{subfigure}%
\begin{subfigure}{.5\textwidth}
  \centering
  \includegraphics[width=\linewidth]{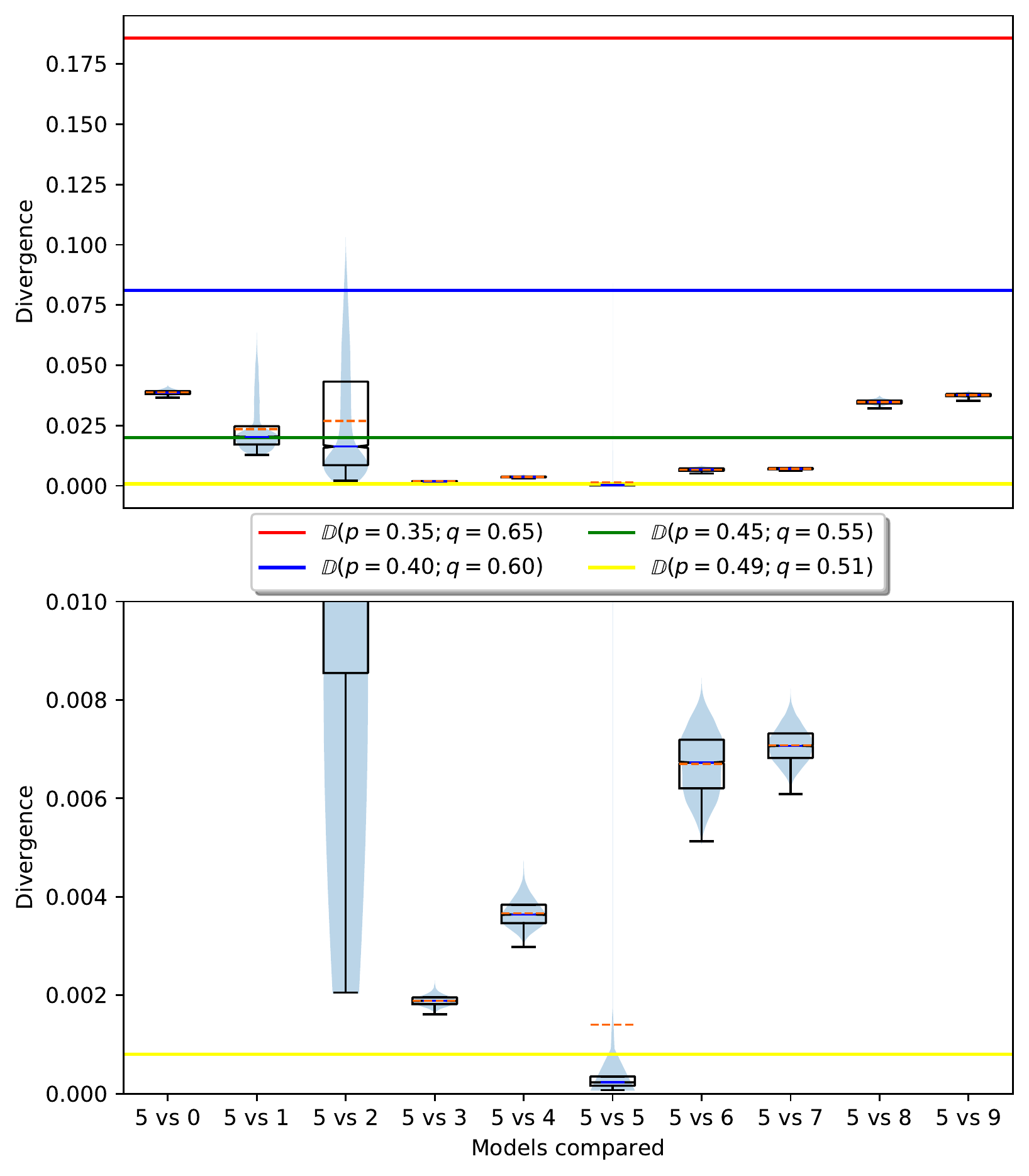}
\end{subfigure}
\begin{subfigure}{.5\textwidth}
  \centering
  \includegraphics[width=\linewidth]{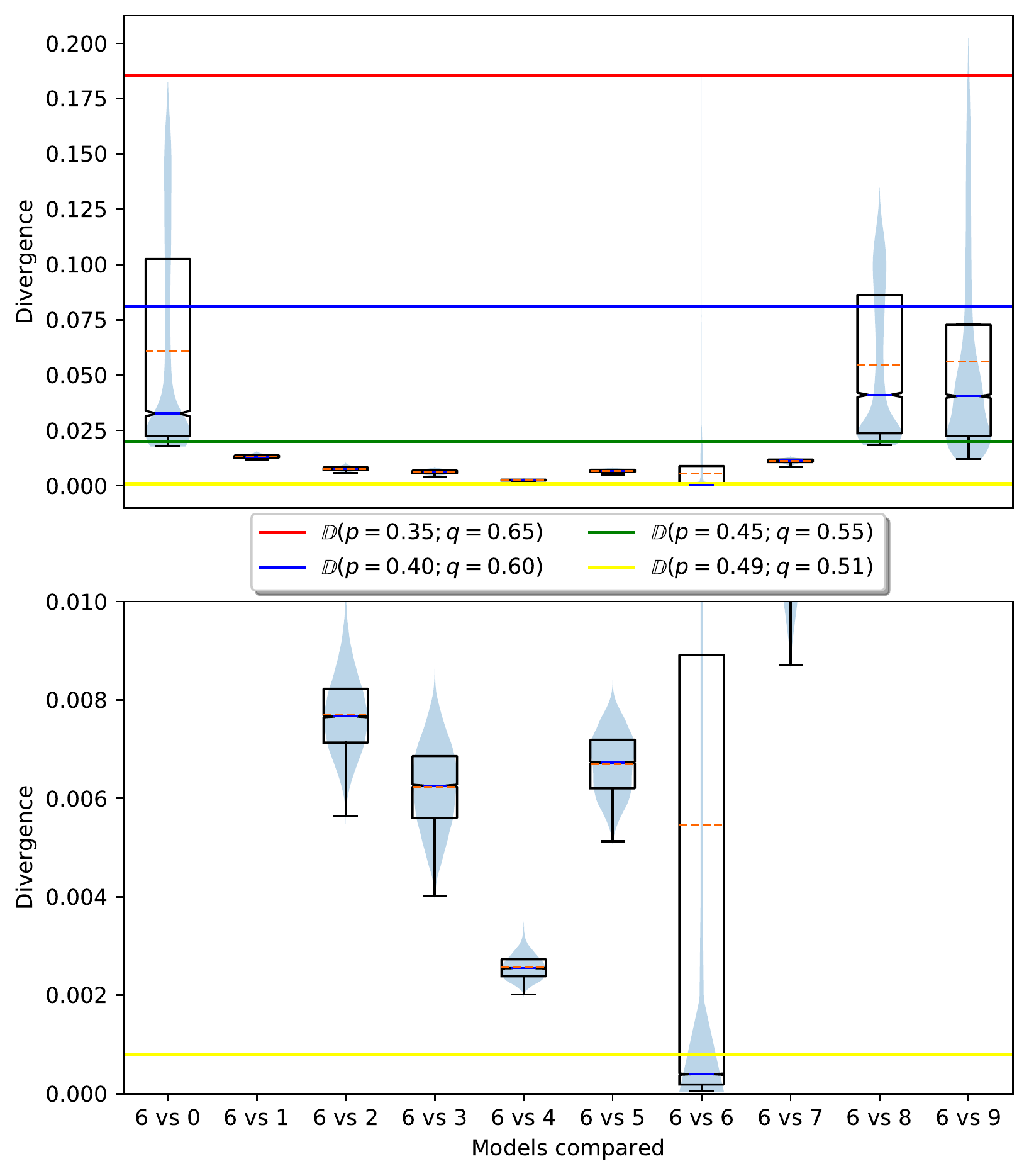}
\end{subfigure}%
\begin{subfigure}{.5\textwidth}
  \centering
  \includegraphics[width=\linewidth]{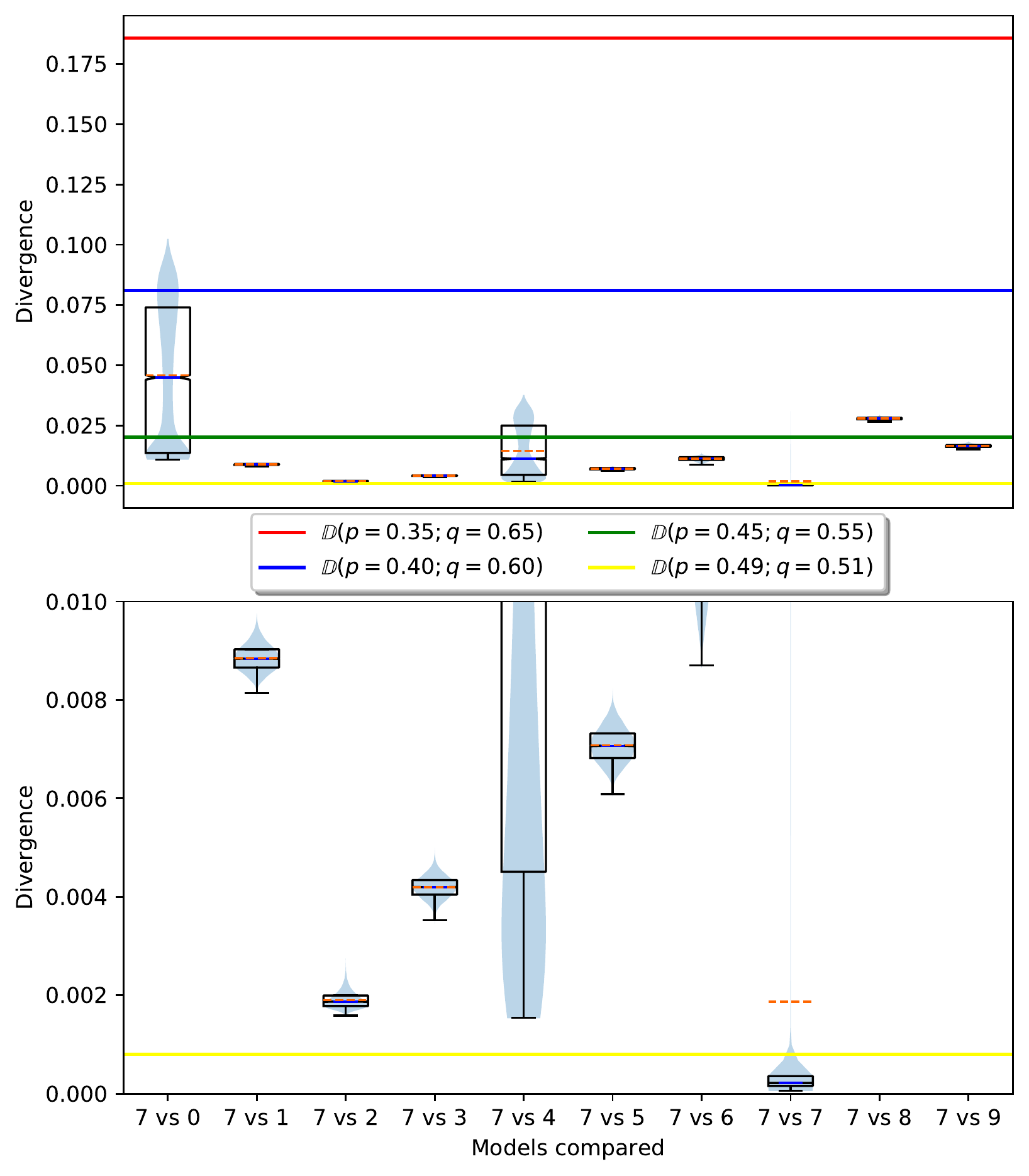}
\end{subfigure}
\caption{Box plots of samples from our divergences comparing categories 4 to 7 to all categories.}
\label{fig:cifar_compare_2}
\end{figure}

\begin{figure}[!htb]
\begin{subfigure}{.5\textwidth}
  \centering
  \includegraphics[width=\linewidth]{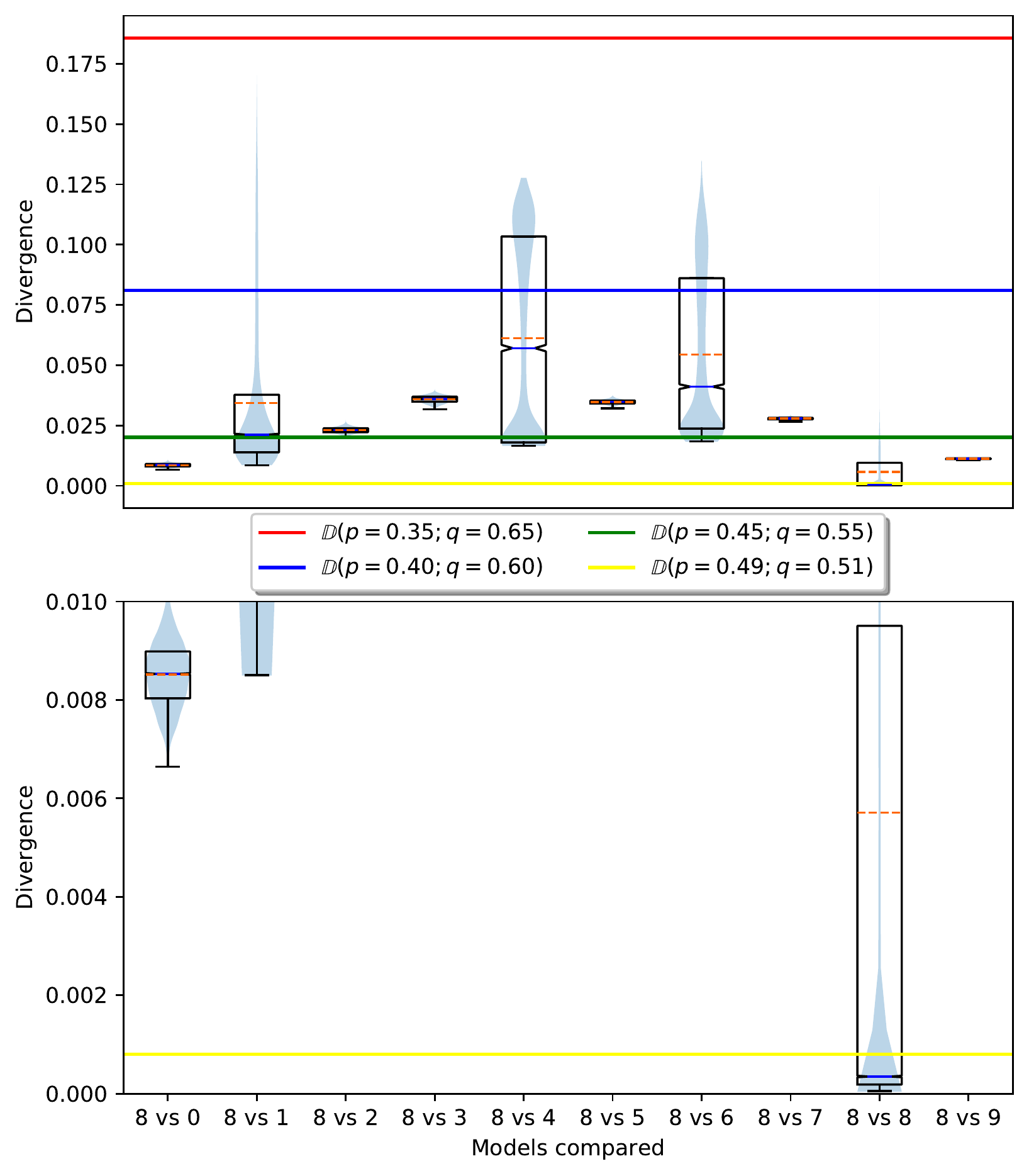}
\end{subfigure}%
\begin{subfigure}{.5\textwidth}
  \centering
  \includegraphics[width=\linewidth]{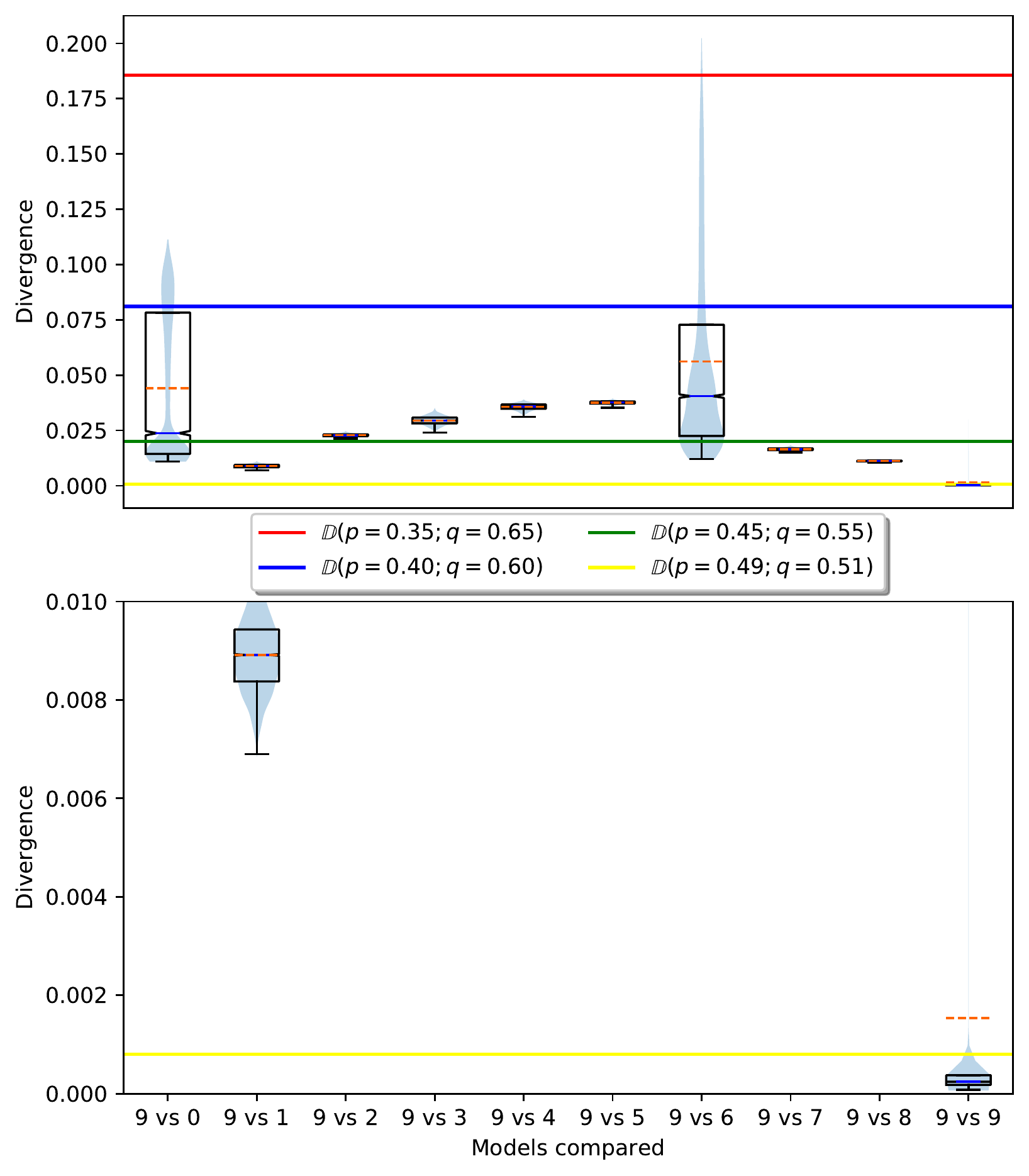}
\end{subfigure}
\caption{Box plots of samples from our divergences comparing categories 8 and 9 to all categories.}
\label{fig:cifar_compare_3}
\end{figure}

In Figures \ref{fig:cifar_compare_1}, \ref{fig:cifar_compare_2} and \ref{fig:cifar_compare_3}, we present the results of such experiment, with box plots of the obtained divergences for all possible category combinations, with the median and mean also plotted for each category and with the divergence of known Bernoulli distributions plotted as horizontal lines.

Except for categories 0, 2 and 4, the samples concentrated all their mass near zero when comparing a category to itself (a desirable behaviour). For categories 0, 2, 4, we have a large mass indicating uncertainty, but even then, there were a considerable amount of samples near zero. Note that the median for these 3 categories is also much closer to zero than the mean, indicating its resilience to outliers.

On the other hand, when making comparisons against distinct categories, we can see cases with high uncertainty (i.e.: box plots with wide extensions; generally with few points close to the zero) to cases with higher certainty and therefore concentrated mass around an specific divergence (i.e.: box plots with narrow extensions).

We conclude that the method is therefore useful for the purpose of data exploration as it works as expected in a complex space such as images (specially since we cannot even assert for sure that images from a single category arise a single data generating process).



\FloatBarrier
\section{Hypothesis testing}
\label{sec:htest}

We can additionally use this method to directly test if two samples come from the same population. One way to do this is to find a threshold value (\textit{cutpoint}) from a decision theoretic stand point where we would reject the null hypothesis of the two samples coming from the same population. This is what is done in \citep{Ceregatti2018} in the case of a
Dirichlet process prior, where the threshold is chosen so as to control type I error of the hypothesis test.
Unfortunately, this is not possible in general and in general the cutpoint depends on the true data generating function.

Given this, we work instead with a simple permutation test where the datasets are repeatedly\footnote{In this work, due to computational limitations we worked with 100 permutations which we set as the default in our software implementation. A higher number of permutations may achieve better stability on the p-value distribution.} permuted against each other (i.e.: their data is mixed), and the average divergence of the samples is used a test statistic. The p-value is then given by the quantile of the non-permuted dataset among all the statistics\footnote{For instance, if 43 of the permuted datasets had resulted on a lower divergence statisticthan that of non-permuted dataset and on the other hand 57 had resulted on greater divergence, then the p-value would be $43/(43+57)=0.43$.}. We note that for the hypothesis test to work in the sense of being a proper test (uniform under the null), it is not necessary to do VAE refits, but however they are expected to increase the test power as we shall see next.
We present the procedure in Algorithm \ref{alg:hypt}.

\begin{algorithm}
 \caption{ \small Obtaning the p-value for hypothesis testing}\label{alg:hypt}
 \textbf{Input:} {\small  
 dataset $D_1$,
 dataset $D_2$,
 number of desired samples per refit $n$,
 number of desired refits $R$,
 number of permutations $t$,
 averaging function $M$ (e.g. mean or median)
 } \\
 \textbf{Output:} {\small  
 p-value $\rho$.
 } 
 \begin{algorithmic}[1]
  \For{$i \in \{0,\ldots,t\}$}
      \State Run Algorithm \ref{alg:comparison}, and store the results in $S_i$.
      \State Calculate $M(S_i)$ and store the result in $K_i$.
      \State Permute the instances of dataset $D_1$.
      \State Permute the instances of dataset $D_2$.
  \EndFor
  \State Obtain the quantile of $K_0$ with regards to $\{K_1, K_2, ..., K_t\}$. Store such quantile in $\rho$. 
 \end{algorithmic}
\end{algorithm}

\subsection{Evaluation (simulated data)}
In this section, we apply the proposed hypothesis testing method to simulated datasets from a known data generating function and plot the observed p-value distribution. The data generating function for the datasets is defined as:
\begin{align*}
\text{lgr}(\mu=log(2), \sigma=\alpha) - \text{lgr}(\mu=log(2), \sigma=\beta) + \text{gr}(\mu=1, \sigma=2) + k
\end{align*}
where lgr stands for multivariate log Gaussian random number generator, and gr stands for multivariate Gaussian random number generator, both with diagonal covariance matrices.
Moreover
$\alpha = {0.2+0.7*i/9}_{i=0}^{i=9}$ and $\beta = {0.5}_{i=0}^{i=9}$

For simplicity, we do not use refits here. The value of the vector $k$ is fixed in zero for one the dataset, and varied for the other, This is done in order to change the dissimilarity between the samples (i.e.: the larger k is, the more dissimilar the sample distributions are) and from that, observe the behaviour of the distribution of the p-value. 

\begin{figure}[!htb]
\begin{subfigure}{.5\textwidth}
  \centering
  \includegraphics[width=0.95\textwidth]{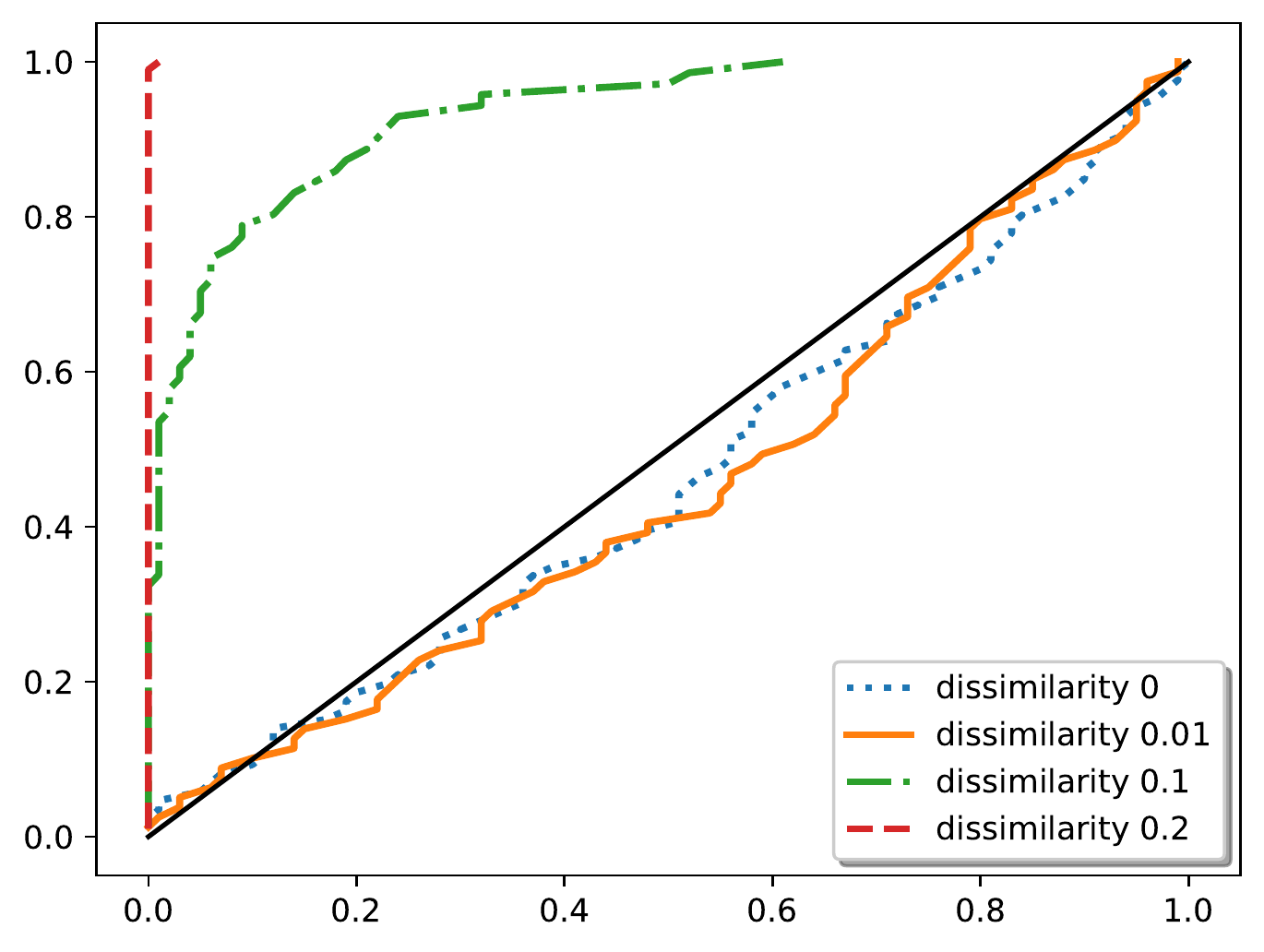}
\caption{Permutation.}
\label{fig:gen_data_htest_permutation}
\end{subfigure}%
\begin{subfigure}{.5\textwidth}
  \centering
  \includegraphics[width=0.95\textwidth]{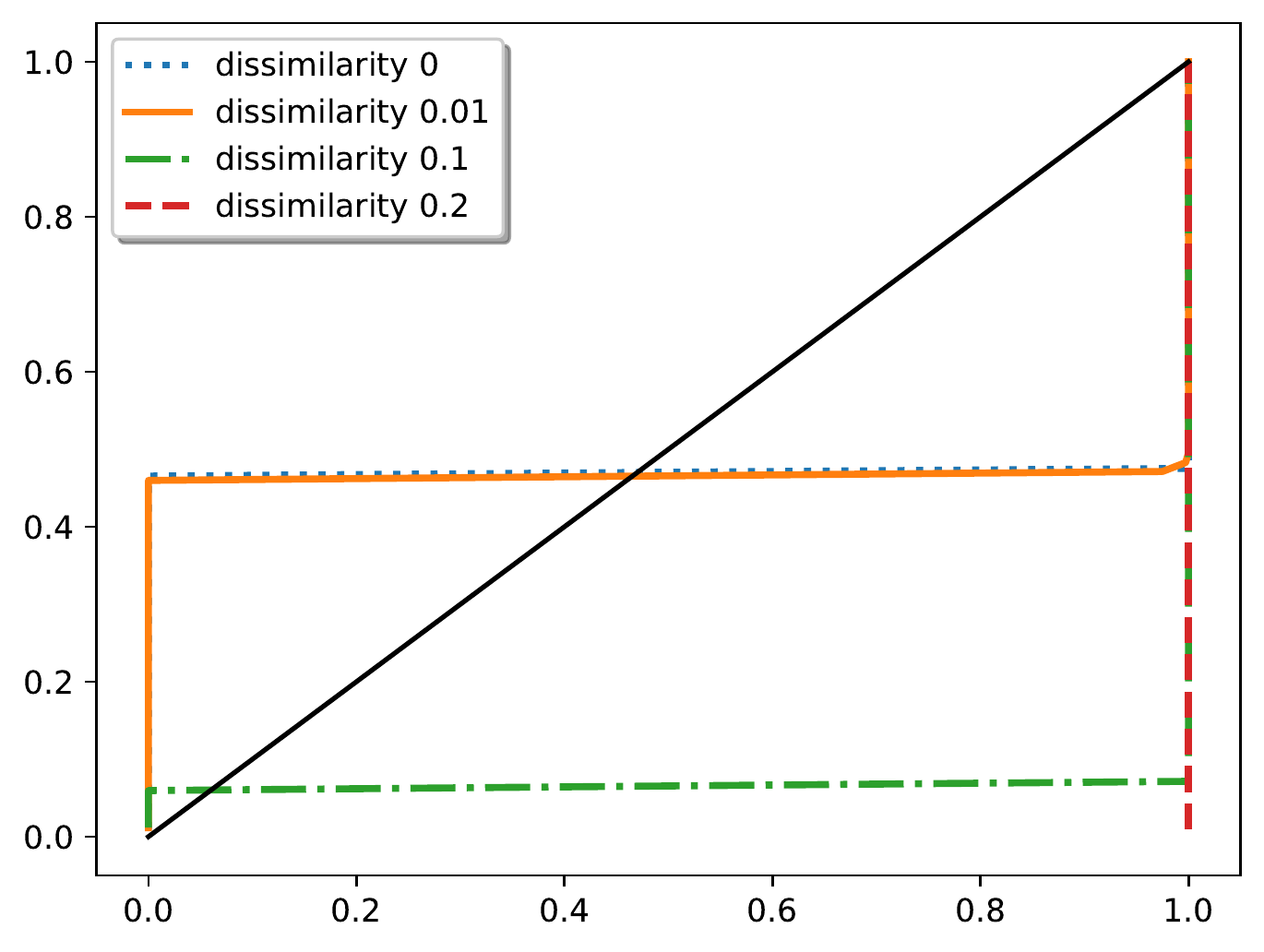}
\caption{Asymptotic.}
\label{fig:gen_data_htest_asymptotic}
\end{subfigure}
\caption{Empirical cumulative distribution function of the p-values for distinct dissimilarity values (when the dissimilarity is zero, the null hypothesis is true) using a permutation test and asymptotic (approximate to permutation test).}
\end{figure}

In Figure \ref{fig:gen_data_htest_permutation}, we present the results of such experiment using the permutation test: the empirical cumulative distribution of the p-values;
while
in Figure \ref{fig:gen_data_htest_asymptotic}, we do the same simulation study using an asymptotic approximate to the permutation test. 
 
The permutation test fulfilled the required properties of a frequentist hypothesis test, as it has approximately (sub)uniform distribution under the null hypothesis (ensured as a property of permutation tests) and the test power increases as the divergence increases (here empirically observable), notice that, for simplicity, we do not use refits here; if we do, we expect the power to increase as is the case in the next section. The asymptotic test on the other hand, performed poorly.

\subsection{Evaluation (images)}

Here, we also applied the hypothesis testing method to the CIFAR10 dataset, using the same 2500 images for each category as described in \ref{subsec:eval_compare_cifar10}.
In Tables
\ref{tab:cifar_htest_pvalues_without_refits_and_median_averaging},
\ref{tab:cifar_htest_pvalues_with_refits_and_median_averaging},
\ref{tab:cifar_htest_pvalues_without_refits_and_mean_averaging} and
\ref{tab:cifar_htest_pvalues_with_refits_and_mean_averaging}
we present the p-values obtained in the test while in Tables
\ref{tab:cifar_htest_errors_without_refits_and_median_averaging},
\ref{tab:cifar_htest_errors_with_refits_and_median_averaging},
\ref{tab:cifar_htest_errors_without_refits_and_mean_averaging} and
\ref{tab:cifar_htest_errors_with_refits_and_mean_averaging}
we present the combinations that gave good results and type 2 error for a significance level of 5\% (there wasn't any case of type 1 error). We applied the tests both without VAE refits and with 5 refits; we also tried the median as an alternative to the mean with the intuition that this might help remove the weight of outlier distance points.

In Table \ref{tab:cifar_htest_summary}, we present a summary of the results.
The method performed well under the null for both the mean and median metrics (with the mean metric presenting better performance). Moreover, the method has shown to have a significant increase in test power when used with VAE refits.

\begin{table}[htbp]
 \centering
 \caption{P-values for hypothesis testing for each categories of the CIFAR10 dataset \textbf{without} refits and averaging using the \textbf{median}. Row/column $c_i$ represents the category $i$.}
\begin{tabular}{lllllllllll}
\toprule
{} &    c0 &    c1 &    c2 &    c3 &    c4 &    c5 &    c6 &    c7 &    c8 &    c9 \\
\midrule
c0 &  0.99 &  0.00 &  0.00 &  0.00 &  0.00 &  0.00 &  0.00 &  0.00 &  0.39 &  0.00 \\
c1 &     - &  0.41 &  0.56 &  0.00 &  0.00 &  0.00 &  0.04 &  0.23 &  0.02 &  0.36 \\
c2 &     - &     - &  0.74 &  0.49 &  0.39 &  0.03 &  0.58 &  0.03 &  0.00 &  0.00 \\
c3 &     - &     - &     - &  0.78 &  0.30 &  0.25 &  0.48 &  0.21 &  0.00 &  0.00 \\
c4 &     - &     - &     - &     - &  0.38 &  0.20 &  0.28 &  0.22 &  0.00 &  0.00 \\
c5 &     - &     - &     - &     - &     - &  0.01 &  0.00 &  0.00 &  0.00 &  0.00 \\
c6 &     - &     - &     - &     - &     - &     - &  0.96 &  0.31 &  0.00 &  0.00 \\
c7 &     - &     - &     - &     - &     - &     - &     - &  0.78 &  0.00 &  0.00 \\
c8 &     - &     - &     - &     - &     - &     - &     - &     - &  0.05 &  0.00 \\
c9 &     - &     - &     - &     - &     - &     - &     - &     - &     - &  0.54 \\
\bottomrule
\end{tabular}

 \label{tab:cifar_htest_pvalues_without_refits_and_median_averaging}
\end{table}

\begin{table}[htbp]
 \centering
 \caption{P-values for hypothesis testing for each categories of the CIFAR10 dataset \textbf{with} refits and averaging using the \textbf{median}. Row/column $c_i$ represents the category $i$.}
\begin{tabular}{lllllllllll}
\toprule
{} &    c0 &    c1 &    c2 &    c3 &    c4 &    c5 &    c6 &    c7 &    c8 &    c9 \\
\midrule
c0 &  0.19 &  0.00 &  0.00 &  0.00 &  0.00 &  0.00 &  0.00 &  0.00 &  0.01 &  0.00 \\
c1 &     - &  0.25 &  0.02 &  0.00 &  0.00 &  0.00 &  0.00 &  0.00 &  0.00 &  0.01 \\
c2 &     - &     - &  0.35 &  0.19 &  0.04 &  0.03 &  0.06 &  0.00 &  0.00 &  0.00 \\
c3 &     - &     - &     - &  0.40 &  0.05 &  0.13 &  0.24 &  0.02 &  0.00 &  0.00 \\
c4 &     - &     - &     - &     - &  0.90 &  0.01 &  0.00 &  0.01 &  0.00 &  0.00 \\
c5 &     - &     - &     - &     - &     - &  0.94 &  0.01 &  0.00 &  0.00 &  0.00 \\
c6 &     - &     - &     - &     - &     - &     - &  0.01 &  0.00 &  0.00 &  0.00 \\
c7 &     - &     - &     - &     - &     - &     - &     - &  1.00 &  0.00 &  0.00 \\
c8 &     - &     - &     - &     - &     - &     - &     - &     - &  0.35 &  0.00 \\
c9 &     - &     - &     - &     - &     - &     - &     - &     - &     - &  0.02 \\
\bottomrule
\end{tabular}

 \label{tab:cifar_htest_pvalues_with_refits_and_median_averaging}
\end{table}

\begin{table}[htbp]
 \centering
 \caption{P-values for hypothesis testing for each categories of the CIFAR10 dataset \textbf{without} refits and averaging using the \textbf{mean}. Row/column $c_i$ represents the category $i$.}
\begin{tabular}{lllllllllll}
\toprule
{} &    c0 &    c1 &    c2 &    c3 &    c4 &    c5 &    c6 &    c7 &    c8 &    c9 \\
\midrule
c0 &  0.24 &  0.11 &  0.00 &  0.00 &  0.00 &  0.00 &  0.00 &  0.00 &  0.47 &  0.04 \\
c1 &     - &  0.82 &  0.50 &  0.22 &  0.02 &  0.04 &  0.06 &  0.14 &  0.00 &  0.04 \\
c2 &     - &     - &  0.18 &  0.48 &  0.47 &  0.43 &  0.08 &  0.01 &  0.09 &  0.00 \\
c3 &     - &     - &     - &  0.21 &  0.10 &  0.35 &  0.14 &  0.26 &  0.00 &  0.00 \\
c4 &     - &     - &     - &     - &  0.87 &  0.19 &  0.39 &  0.00 &  0.00 &  0.00 \\
c5 &     - &     - &     - &     - &     - &  0.18 &  0.04 &  0.22 &  0.00 &  0.00 \\
c6 &     - &     - &     - &     - &     - &     - &  0.73 &  0.00 &  0.00 &  0.00 \\
c7 &     - &     - &     - &     - &     - &     - &     - &  0.47 &  0.00 &  0.01 \\
c8 &     - &     - &     - &     - &     - &     - &     - &     - &  0.44 &  0.05 \\
c9 &     - &     - &     - &     - &     - &     - &     - &     - &     - &  0.47 \\
\bottomrule
\end{tabular}

 \label{tab:cifar_htest_pvalues_without_refits_and_mean_averaging}
\end{table}

\begin{table}[htbp]
 \centering
 \caption{P-values for hypothesis testing for each categories of the CIFAR10 dataset \textbf{with} refits and averaging using the \textbf{mean}. Row/column $c_i$ represents the category $i$.}
\begin{tabular}{lllllllllll}
\toprule
{} &    c0 &    c1 &    c2 &    c3 &    c4 &    c5 &    c6 &    c7 &    c8 &    c9 \\
\midrule
c0 &  0.15 &  0.00 &  0.00 &  0.00 &  0.00 &  0.00 &  0.00 &  0.00 &  0.02 &  0.00 \\
c1 &     - &  0.29 &  0.06 &  0.00 &  0.00 &  0.00 &  0.00 &  0.00 &  0.00 &  0.01 \\
c2 &     - &     - &  0.09 &  0.05 &  0.11 &  0.36 &  0.03 &  0.00 &  0.00 &  0.00 \\
c3 &     - &     - &     - &  0.83 &  0.07 &  0.44 &  0.14 &  0.01 &  0.00 &  0.00 \\
c4 &     - &     - &     - &     - &  0.53 &  0.04 &  0.09 &  0.19 &  0.00 &  0.00 \\
c5 &     - &     - &     - &     - &     - &  0.50 &  0.00 &  0.00 &  0.00 &  0.00 \\
c6 &     - &     - &     - &     - &     - &     - &  0.11 &  0.04 &  0.00 &  0.00 \\
c7 &     - &     - &     - &     - &     - &     - &     - &  0.47 &  0.00 &  0.00 \\
c8 &     - &     - &     - &     - &     - &     - &     - &     - &  0.17 &  0.00 \\
c9 &     - &     - &     - &     - &     - &     - &     - &     - &     - &  0.71 \\
\bottomrule
\end{tabular}

 \label{tab:cifar_htest_pvalues_with_refits_and_mean_averaging}
\end{table}

\begin{table}[htbp]
 \centering
 \caption{Results of the hypothesis testing when applying a critical rate of 5\% \textbf{without} refits and averaging using the \textbf{median}. Here G stands for ``good'' and E2 for type 2 error.}
\begin{tabular}{lllllllllll}
\toprule
{} & c0 & c1 &  c2 &  c3 &  c4 &  c5 &  c6 &  c7 &  c8 &  c9 \\
\midrule
c0 &  G &  G &   G &   G &   G &   G &   G &   G &  E2 &   G \\
c1 &  - &  G &  E2 &   G &   G &   G &   G &  E2 &   G &  E2 \\
c2 &  - &  - &   G &  E2 &  E2 &   G &  E2 &   G &   G &   G \\
c3 &  - &  - &   - &   G &  E2 &  E2 &  E2 &  E2 &   G &   G \\
c4 &  - &  - &   - &   - &   G &  E2 &  E2 &  E2 &   G &   G \\
c5 &  - &  - &   - &   - &   - &  E1 &   G &   G &   G &   G \\
c6 &  - &  - &   - &   - &   - &   - &   G &  E2 &   G &   G \\
c7 &  - &  - &   - &   - &   - &   - &   - &   G &   G &   G \\
c8 &  - &  - &   - &   - &   - &   - &   - &   - &  E1 &   G \\
c9 &  - &  - &   - &   - &   - &   - &   - &   - &   - &   G \\
\bottomrule
\end{tabular}

 \label{tab:cifar_htest_errors_without_refits_and_median_averaging}
\end{table}

\begin{table}[htbp]
 \centering
 \caption{Results of the hypothesis testing when applying a critical rate of 5\% \textbf{with} refits and averaging using the \textbf{median}. Here G stands for ``good'' and E2 for type 2 error.}
\begin{tabular}{lllllllllll}
\toprule
{} & c0 & c1 & c2 &  c3 & c4 &  c5 &  c6 & c7 & c8 &  c9 \\
\midrule
c0 &  G &  G &  G &   G &  G &   G &   G &  G &  G &   G \\
c1 &  - &  G &  G &   G &  G &   G &   G &  G &  G &   G \\
c2 &  - &  - &  G &  E2 &  G &   G &  E2 &  G &  G &   G \\
c3 &  - &  - &  - &   G &  G &  E2 &  E2 &  G &  G &   G \\
c4 &  - &  - &  - &   - &  G &   G &   G &  G &  G &   G \\
c5 &  - &  - &  - &   - &  - &   G &   G &  G &  G &   G \\
c6 &  - &  - &  - &   - &  - &   - &  E1 &  G &  G &   G \\
c7 &  - &  - &  - &   - &  - &   - &   - &  G &  G &   G \\
c8 &  - &  - &  - &   - &  - &   - &   - &  - &  G &   G \\
c9 &  - &  - &  - &   - &  - &   - &   - &  - &  - &  E1 \\
\bottomrule
\end{tabular}

 \label{tab:cifar_htest_errors_with_refits_and_median_averaging}
\end{table}

\begin{table}[htbp]
 \centering
 \caption{Results of the hypothesis testing when applying a critical rate of 5\% \textbf{without} refits and averaging using the \textbf{mean}. Here G stands for ``good'' and E2 for type 2 error.}
\begin{tabular}{lllllllllll}
\toprule
{} & c0 &  c1 &  c2 &  c3 &  c4 &  c5 &  c6 &  c7 &  c8 & c9 \\
\midrule
c0 &  G &  E2 &   G &   G &   G &   G &   G &   G &  E2 &  G \\
c1 &  - &   G &  E2 &  E2 &   G &   G &  E2 &  E2 &   G &  G \\
c2 &  - &   - &   G &  E2 &  E2 &  E2 &  E2 &   G &  E2 &  G \\
c3 &  - &   - &   - &   G &  E2 &  E2 &  E2 &  E2 &   G &  G \\
c4 &  - &   - &   - &   - &   G &  E2 &  E2 &   G &   G &  G \\
c5 &  - &   - &   - &   - &   - &   G &   G &  E2 &   G &  G \\
c6 &  - &   - &   - &   - &   - &   - &   G &   G &   G &  G \\
c7 &  - &   - &   - &   - &   - &   - &   - &   G &   G &  G \\
c8 &  - &   - &   - &   - &   - &   - &   - &   - &   G &  G \\
c9 &  - &   - &   - &   - &   - &   - &   - &   - &   - &  G \\
\bottomrule
\end{tabular}

 \label{tab:cifar_htest_errors_without_refits_and_mean_averaging}
\end{table}

\begin{table}[htbp]
 \centering
 \caption{Results of the hypothesis testing when applying a critical rate of 5\% \textbf{with} refits and averaging using the \textbf{mean}. Here G stands for ``good'' and E2 for type 2 error.}
\begin{tabular}{lllllllllll}
\toprule
{} & c0 & c1 &  c2 & c3 &  c4 &  c5 &  c6 &  c7 & c8 & c9 \\
\midrule
c0 &  G &  G &   G &  G &   G &   G &   G &   G &  G &  G \\
c1 &  - &  G &  E2 &  G &   G &   G &   G &   G &  G &  G \\
c2 &  - &  - &   G &  G &  E2 &  E2 &   G &   G &  G &  G \\
c3 &  - &  - &   - &  G &  E2 &  E2 &  E2 &   G &  G &  G \\
c4 &  - &  - &   - &  - &   G &   G &  E2 &  E2 &  G &  G \\
c5 &  - &  - &   - &  - &   - &   G &   G &   G &  G &  G \\
c6 &  - &  - &   - &  - &   - &   - &   G &   G &  G &  G \\
c7 &  - &  - &   - &  - &   - &   - &   - &   G &  G &  G \\
c8 &  - &  - &   - &  - &   - &   - &   - &   - &  G &  G \\
c9 &  - &  - &   - &  - &   - &   - &   - &   - &  - &  G \\
\bottomrule
\end{tabular}

 \label{tab:cifar_htest_errors_with_refits_and_mean_averaging}
\end{table}

\begin{table}[htbp]
 \centering
 \caption{Summary of the results of the hypothesis testing when applying a critical rate of 5\%.}

\begin{tabular}{llll}
\toprule
Averaging &   Refits & Number Type I errors & Number Type II errors \\
\midrule
     mean &     with &                    0 &                    16 \\
     mean &  without &                    0 &                    36 \\
   median &     with &                    2 &                     8 \\
   median &  without &                    2 &                    30 \\
\bottomrule
\end{tabular}

 \label{tab:cifar_htest_summary}
\end{table}

\FloatBarrier

\section{Discussion and Conclusions}
\label{sec:conclusion}

In this work, we proposed and applied a novel method of two sample distance measurement and hypothesis testing to simulated and real-world datasets. 
The proposed methods could be used for various tasks in the machine learning pipeline, including: 
\begin{itemize}
    \item Distribution shift detection and measurement: a dataset from a experiment done in one month (e.g.: opinions of customers on a product on a specific month) might diverge in distribution from a dataset collected in another month. With our method it's possible to measure and test this diverge. 
    \item Dataset split: to address overfitting, the data is usually split into train, validation and/or test parts. To be able to develop robust models, these parts should be similar, but should also differ enough to ensure generalization. With the proposed methods the dataset split can be done in a controlled manner. 
    \item Self-supervised clustering: based on the distance, by fine-tuning the threshold (cutpoint), binary or multi-class clustering could be performed.
    \item Anomaly detection: applying the proposed method to processes where anomaly may occur (e.g. malicious attack, malfunction, etc.). In this case, the distance measurement can give a direct feedback of how much the actual behaviour differs from the normal one.
\end{itemize}

We concluded that both two sample distance measurement and hypothesis testing were able to satisfactorily perform the intended tasks on the tested simulated and real world datasets. 

\section*{Acknowledgments}
\label{sec:acknowledgements}
Marco In\'{a}cio is grateful for the financial support of CAPES (this study was financed in part by the Coordena\c{c}\~{a}o de Aperfei\c{c}oamento de Pessoal de N\'{\i}vel Superior - Brasil (CAPES) - Finance Code 001) and of the Erasmus+ program.
Rafael Izbicki is grateful for the financial support of FAPESP (grants 2017/03363-8 and 2019/11321-9) and CNPq (grant 306943/2017-4).
B\'alint Gyires-T\'oth is grateful for the financial support of the BME-Artificial Intelligence FIKP grant of Ministry of Human Resources (BME FIKP-MI/SC) and for the financial support of Doctoral Research Scholarship of Ministry of Human Resources (\'UNKP-19-4-BME-189) in the scope of New National Excellence Program, by J\'anos Bolyai Research Scholarship of the Hungarian Academy of Sciences.
The authors are also grateful for the suggestions
given by Rafael Bassi Stern.

\printbibliography

\appendix
\section*{Appendix: Neural networks configuration, software and package}
\label{sec:configs_and_software}
We work with a dense neural network of 10 layers with 100 neurons on each layer (totaling 195060 parameters), for both encoder and decoder networks, and the following additional specification:

\begin{itemize}
\item \textbf{Optimizer}: we work with the Adamax optimizer \citep{adam-optim} with initial learning rate of $0.01$ and decrease its learning rate by half if improvement is seen on the validation loss for a considerable number of epochs.

\item \textbf{Initialization}: we used the initialization method proposed by \citep{nn-initialization}.

\item \textbf{Layer activation}: we chose ELU \citep{elu} as activation functions.

\item \textbf{Stop criterion}: a 90\%/10\% split early stopping for small datasets and a higher split factor for larger datasets (increasing the proportion of training instances) and a patience of 50 epochs without improvement on the validation set.

\item \textbf{Normalization and number of hidden layers}: batch normalization, as proposed by \citep{batch-normalization}, is used in this work in order to speed-up the training process.

\item \textbf{Dropout}: here we also make use of dropout which as proposed by \citep{dropout} (with dropout rate of 0.5).

\item \textbf{Software}: we have PyTorch as framework of choice which works with automatic differentiation and the software implementation of work is available at \url{https://github.com/randommm/vaecompare}.
\end{itemize}

\end{document}